\documentclass[letterpaper, 10pt, conference]{ieeeconf}
\IEEEoverridecommandlockouts
\usepackage{cite}
\usepackage{amsmath,amssymb,amsfonts}
\usepackage{graphicx}
\usepackage{textcomp}
\usepackage{xcolor}
\usepackage{booktabs}
\usepackage{multirow}
\usepackage{array}
\newcolumntype{L}[1]{>{\raggedright\arraybackslash}p{#1}}
\usepackage{url}
\usepackage[hidelinks]{hyperref}
\usepackage[capitalize]{cleveref}
\newcommand{\CFR}{\mathrm{CFR}}
\newcommand{\HS}{\mathrm{HS}}
\newcommand{\SP}{\mathrm{SP}}
\newcommand{\NumCFRlo}{0.23}
\newcommand{\NumCFRhi}{0.47}

\newcommand{\NumHSlo}{0.00}
\newcommand{\NumHShi}{0.10}

\newcommand{\NumDetNight}{89.9}
\newcommand{\NumDetStill}{8.5}

\newcommand{\NumDetOrig}{100}

\newcommand{\NumMapShare}{95}

\newcommand{\NumHumanRef}{0.57}
\newcommand{\NumHumanSc}{0.13}

\newcommand{\NumScCFR}{3}
\newcommand{\NumScExp}{13}

\newcommand{\NumPool}{170}
\newcommand{\NumCollGap}{0.4}
\newcommand{\NumCollGapEight}{3.9}
\newcommand{\NumNstarCFRhi}{14}
\newcommand{\NumNstarExpHi}{51}
\newcommand{\NumRankSPforty}{99}
\newcommand{\NumRankExpTen}{96}
\newcommand{\NumRankHSninety}{84}

\newcommand{\NumPass}{4}
\newcommand{\NumNightDD}{37}
\newcommand{\NumNightLTF}{12}
\newcommand{\NumNightSL}{16}
\newcommand{\NumNightDDv}{-8}
\newcommand{\NumRhoTTC}{0.77}

\newcommand{\NumRhoTTCExpL}{1.00}
\newcommand{\NumRhoTTCExpR}{0.77}

\newcommand{\NumMaxdLtwo}{0.07}
\newcommand{\NumMaxdCol}{1.7}
\newcommand{\NumMaxdPed}{3.0}
\newcommand{\NumRhoLtwoExp}{0.66}
\newcommand{\NumRhoAllClose}{-0.09}

\newcommand{\NumNclose}{246}

\newcommand{\NumSGscenes}{246}

\newcommand{\NumFgtIlo}{4}
\newcommand{\NumFgtIhi}{23}
\newcommand{\NumRhoLo}{$-$0.37}
\newcommand{\NumRhoHi}{0.07}
\newcommand{\NumMoveN}{121}
\newcommand{\NumMoveFar}{43}
\newcommand{\NumMoveNear}{42}

\newcommand{\NumSLnear}{32}
\newcommand{\NumSLfar}{13}
\newcommand{\NumSLmed}{$-$0.59}

\newcommand{\NumDDmed}{2.16}

\newcommand{\NumSpSL}{0.44}
\newcommand{\NumSpOtherLo}{0.65}
\newcommand{\NumSpOtherHi}{0.84}
\newcommand{\NumScalLo}{$-$0.28}
\newcommand{\NumScalHi}{0.23}
\newcommand{\NumScalNeg}{5}
\newcommand{\NumScalWorst}{DiffusionDrive}
\newcommand{\NumScalTtcLo}{$-$0.07}
\newcommand{\NumScalTtcHi}{0.32}

\newcommand{\NumCfModels}{6}
\newcommand{\NumSpBehTab}{0.83}
\newcommand{\NumExpBehTab}{{-}0.77}
\newcommand{\NumCfrBehTab}{{-}0.37}
\newcommand{\NumRhoLtwoKeep}{0.43}
\newcommand{\NumRhoLbKeep}{0.09}
\newcommand{\NumRhoPedcolKeep}{0.81}

\newcommand{\NumStdRhoMax}{0.43}
\newcommand{\NumOtherSideTTC}{0.77}
\newcommand{\NumSpLhdList}{DD 0.87, LTF 0.78, DDv2 0.66, SimLingo 0.37, AutoVLA 0.79, Alpamayo 1.5 0.85}
\newcommand{\NumCfCells}{4428}
\newcommand{\NumCfNeed}{787}
\newcommand{\NumCfReal}{15}
\newcommand{\NumCfRealPct}{1.9}

\newcommand{\NumCfNeedLo}{11}
\newcommand{\NumCfNeedHi}{30}
\newcommand{\NumCfDsLo}{0.00}
\newcommand{\NumCfDsHi}{0.03}
\newcommand{\NumCfArcLo}{$-$0.02}
\newcommand{\NumCfArcHi}{0.08}

\begin{document}
\title{\LARGE \bf Beyond the Leaderboard: Counterfactual Diagnosis of End-to-End and VLA Driving Policies Under Domain Shift}
\newif\ifanon \anonfalse
\ifanon
\author{Anonymous authors\\ Paper ID ---}
\else
\author{Ruolin Yang$^{1}$, Zilin Huang$^{1}$, Buoyue Wang$^{1}$, Zhengyang Wan$^{1}$, Yuhao Luo$^{1}$, Zihao Sheng$^{1}$, and Sikai Chen$^{1,*}$%
\thanks{This research was supported by the U.S. Department of Transportation (No.\ 69A3552348305) and by grants from NVIDIA utilizing NVIDIA RTX PRO 6000 Blackwell Max-Q Workstation Edition GPUs.}%
\thanks{$^{1}$Department of Civil and Environmental Engineering, University of Wisconsin--Madison, Madison, WI 53706, USA.}%
\thanks{$^{*}$Corresponding author: Sikai Chen.}%
}
\fi
\maketitle
\thispagestyle{empty}
\pagestyle{empty}
\begin{abstract}
End-to-end and vision-language-action (VLA) driving policies are compared by leaderboard rank, but a rank reports an outcome, not the behaviour behind it, so it predicts poorly how a policy will behave at a new site.
On six released policies, rank on nuScenes open-loop error or on NAVSIM's leaderboard does not carry over to scenes with a pedestrian near the ego corridor at a new site.
We propose a counterfactual check-up: a few hundred real frames, each edited two ways (pedestrian removed, or re-lit by a night-style perturbation), every edit verified by an independent detector, and the change in the planned trajectory read as a diagnosis rather than a score. From these edits two causal axes are read, and five exams built on them separate what a score merges: how far the policy plans to drive, whether seeing the pedestrian buys safety, whether that response scales with danger, whether the plan moves when nothing requires it, and how much an irrelevant lighting change moves it.
On \NumSGscenes{} NAVSIM near-pedestrian scenes, in the cells where the pedestrian lies on the planned path only \NumCfRealPct\% of responses are genuine avoidance, and under our open-loop protocol the median clearance change is at most \NumCfDsHi\,m and the median change in planned distance at most \NumCfArcHi\,m for every policy. In a pre-registered test from left- to right-hand drive, the exposure and specificity orderings, the lighting verdict and the collision outcome transfer, while point values and the hazard-sensitivity verdict do not. Read as a selection report, the profiles say which policy is safe because it plans short, which covers a human-like distance without yielding, and which is unsteady under a change that requires no reaction, and they price each verdict: most settle within a few dozen frames, hazard sensitivity needs hundreds. Code and edited frames will be released.
\end{abstract}

\section{Introduction}
\label{sec:intro}
A leaderboard rank does not tell you how a driving policy will behave in a new domain. End-to-end and vision-language-action (VLA) driving policies are compared by benchmark scores, and a deployment decision is, implicitly, a bet that the rank those scores produce will still hold once the policy is driving at the new site, not just on someone else's roads.
On scenes where a pedestrian is near the ego corridor, that bet fails for the six released policies we test: their rank on nuScenes'~\cite{caesar2020nuscenes} open-loop error or on NAVSIM's~\cite{dauner2024navsim} leaderboard score does not carry over to the deployment site (rank correlation $|\rho|\le\NumStdRhoMax$ against near-pedestrian behaviour there, \cref{sec:sample}), a gap that open-loop scores are known to open~\cite{zhao2026bridgesim,li2024egostatus,codevilla2019exploring}. The reason is not noise but mechanism. A score is an outcome measure, and an outcome can be right for the wrong reason: a policy that never approaches a pedestrian because it plans short trajectories scores as well as one that sees the pedestrian and yields, and the two behave very differently the first time the scene changes. Telling them apart means asking why the score came out as it did, usually from little data, because a closed-loop simulator for the target domain or a fleet test is exactly what the decision is meant to save.

Deployment safety cases still rest on target-domain miles~\cite{kusano2025waymo56m}, and post-training on a policy's own deployments~\cite{pi2025pistar06} presumes the same access. For a selection decision, that cost is prohibitive~\cite{kalra2016driving}.

Cross-domain work in domain adaptation~\cite{ganin2015dann} and generalisation~\cite{gulrajani2021lost}
solves the shift through training and does not assess how a policy behaves in the target domain, so
it leaves the selection problem at deployment open.

Closer to our setting, \cite{kachaev2025don} improves a VLA's out-of-distribution generalisation by
aligning its visual representations, but it presumes an already-chosen vision--language model and
offers no procedure for selecting among policies, vision-only ones included. Representation engineering~\cite{zou2023repe} locates and steers concepts in a language model's activations, but it is built around text prompts. Function
vectors~\cite{todd2024functionvectorslargelanguage} show that a task can be read out of, and
transferred through, a language model's hidden states. We know of no counterpart of these instruments for the cross-domain evaluation of driving policies.

\begin{figure*}[!t]
\centering
\includegraphics[width=\textwidth]{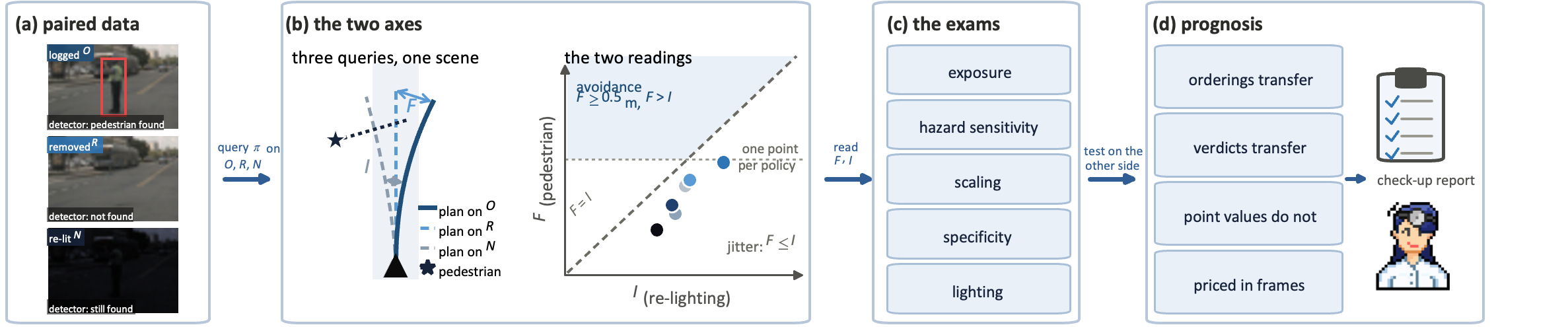}

\caption{\textbf{The Check-Up framework.} Pixel-level paired data (a) yield the two causal axes (b) and the five exams built on them (c). Diagnosed on one driving side and tested on the other, they give a priced prognosis, the check-up report (d).}

\label{fig:method}
\end{figure*}
To address this, we propose the \emph{Check-Up}, a counterfactual diagnosis in four stages (\cref{fig:method}). (a)~We construct pixel-level paired data: every logged frame is queried with and
without the target pedestrian and under a night-style photometric perturbation, and an
independent detector verifies every edit. (b)~From each pair we distil two axes of causal
attribution---the \emph{faithfulness} displacement $F$, the plan change caused by the pedestrian
itself, and the \emph{interference} displacement $I$, the change the same policy produces under an
edit that carries no driving-relevant information~\eqref{eq:FI}. (c)~Five exams, each with a control and a reference threshold, are built on these paired plans and axes (\cref{fig:method}c): how far the policy plans
to drive, whether seeing the pedestrian buys safety, whether that benefit scales with the hazard,
whether the plan moves when nothing requires it, and how much the irrelevant edit moves it. (d)~Read on one driving side and tested on the other, the exams yield a prognosis for the target domain: which verdicts and orderings carry over, what each verdict costs in frames, and a per-policy selection report.

The contribution is the Check-Up, a counterfactual diagnosis of end-to-end and VLA driving
policies under domain shift, in three parts.
\begin{enumerate}\itemsep2pt
\item \emph{Pixel-level paired data.} Every logged frame is queried with and without the target
      pedestrian and under a night-style perturbation, each edit verified by an
      independent detector. Six released policies are read on two driving sides.
\item \emph{Two causal axes.} A response is read as the faithfulness displacement $F$ and the
      interference displacement $I$~\eqref{eq:FI}, and counts as avoidance only if it moves and buys distance, and exceeds the interference where a re-lit plan exists. Where the pedestrian lies on the planned path, \NumCfRealPct\%
      of responses qualify, and the lighting edit moves every plan further than the pedestrian does.
\item \emph{A priced prognosis.} Diagnosed on one driving side and tested on the other, the exposure and specificity orderings, the lighting verdict and the collision outcome transfer, while point values and the hazard-sensitivity verdict do not. Each verdict is priced in frames, lighting within \NumNstarCFRhi, exposure within \NumNstarExpHi, and the profiles read as a selection report that says what each policy needs before deployment (\cref{sec:sample}).
\end{enumerate}

\section{Related Work}
\label{sec:related}

\paragraph{Driving policies and their benchmarks}
We examine bird's-eye-view (BEV) planners of the TransFuser lineage~\cite{chitta2023transfuser} (LTF,
DiffusionDrive~\cite{liao2025diffusiondrive}, DiffusionDriveV2~\cite{diffusiondrivev2}) and
vision--language--action models (SimLingo~\cite{renz2025simlingo}, AutoVLA~\cite{zhou2025autovla},
Alpamayo-1.5~\cite{nvidia2025alpamayo,nvidia2026alpamayo15}). NAVSIM~\cite{dauner2024navsim} scores
plans by multiplying hard gates with a weighted average of soft terms, open-loop scores are known to reward proxies~\cite{li2024egostatus}, end-to-end policies to exploit biases in their inputs~\cite{jaeger2023hidden}, and closed-loop benchmarks~\cite{jia2024bench2drive} score a policy in simulation. NAVSIM's pseudo-simulation~\cite{cao2025pseudosim} moves the score toward closed loop. We ask a different question, whether the pedestrian entered the plan, and read released policies on real frames.

\paragraph{Behavioural tests and causal confusion}
Checklists~\cite{ribeiro2020checklist} and corruption benchmarks~\cite{hendrycks2019benchmarking}
probe models beyond aggregate accuracy, but a metric drop does not say whether a policy became more
cautious or less. Imitation learners respond to correlates of the expert's action rather than to its
causes~\cite{dehaan2019causal,codevilla2019exploring,geirhos2020shortcut}. Our exams intervene on a
cause (a pedestrian) and a nuisance (lighting) directly and read signed plan changes.

\paragraph{Counterfactual editing}
Counterfactual explanations edit the scene~\cite{jacob2022steex}, and
CausalAgents~\cite{roelofs2022causalagents} deletes non-causal agents to test motion forecasters. Closed-loop simulators insert or re-render agents in reconstructed scenes~\cite{ljungbergh2024neuroncap,yang2023unisim,zhou2024hugsim}. Our removal edit applies the same idea at pixel level to real frames, every edit is verified by a detector, and six released policies are compared on the same edits.

\section{The Check-Up: Interventions, Axes and Exams}
\label{sec:checkup}

\subsection{Units, interventions and plans}
Write a policy as a map $\pi$ from a sensory-and-state input to a planned trajectory
$X=(X_t)_{t\in[0,2.5]}$, $X_t\in\mathbb{R}^2$ in the ego frame, sampled at $0.1$\,s. A \emph{unit} is a
pair $u=(f,v)$: a logged frame $f$ and an input ego speed $v$. We take $v$ from the log for every
frame and, wherever the pedestrian's logged future is available, additionally
$v\in\{4,8\}\,\mathrm{m/s}$, which places the pedestrian on the planned path in scenes where the logged
ego was stationary. The same grid is used on both corpora so that all six policies are read on one
unit pool. The readouts are insensitive to it (widening the grid to $\{2,4,6,8\}$ leaves the
specificity and scaling orderings identical, $\rho=1.00$).

Three interventions act on the input, leaving the policy untouched:
\begin{equation}
\begin{aligned}
&\mathcal{E}_O=\mathrm{id},\qquad \mathcal{E}_R:\ \text{delete the pedestrian},\\
&\mathcal{E}_N:\ \text{re-light at night},
\end{aligned}
\label{eq:interventions}
\end{equation}
and yield the paired plans $X^{O}=\pi(\mathcal{E}_O u)$, $X^{R}=\pi(\mathcal{E}_R u)$,
$X^{N}=\pi(\mathcal{E}_N u)$. The letters $O,R,N$ name the three conditions only: an intervention carries one as a subscript, a plan as a superscript. All quantities below are functionals of
these plans and of the
pedestrian's \emph{logged} future $P_t$. The horizon $2.5$\,s is the one all six policies share.

The displacement between two plans is
\begin{equation}
\bar D(X,Y)=\tfrac{1}{|\mathcal{T}|}\textstyle\sum_{t\in\mathcal{T}}\|X_t-Y_t\|,
\label{eq:disp}
\end{equation}
with $\mathcal{T}=\{0,0.1,\dots,2.5\}$, and the two quantities the rest of the paper is written in are
\begin{equation}
F=\bar D(X^{O},X^{R}),\qquad I=\bar D(X^{O},X^{N}).
\label{eq:FI}
\end{equation}
We call $F$ the \emph{faithfulness} displacement---the plan change attributable to the pedestrian
itself---and $I$ the \emph{interference} displacement---the change the same policy produces under a large edit that should not change the correct plan, which bounds from above how far the plan moves for reasons unrelated to the pedestrian. A policy that
uses the pedestrian and ignores the lighting has $F\gg I$. The reverse ordering means the plan is
driven by something the road does not care about.

\paragraph{Realising $\mathcal{E}_R$}
The target is localised by projecting its annotated 3D box into the camera rather than by taking the
dataset's 2D boxes, whose association to the tracked object is unreliable for the short-range,
partially occluded pedestrians of these scenes. It is segmented with SAM~\cite{kirillov2023sam} (\cref{fig:method}a) (box
and torso prompts, clipped to the box, dilated by $9$\,px) and inpainted with
LaMa~\cite{suvorov2022lama}. For the LiDAR policy the points inside the box are deleted. VLA policies read several cameras and only the front one is edited, so we retain only frames in which the target
appears in no other view.

\paragraph{Realising $\mathcal{E}_N$}
Each image $\mathbf{f}\in[0,1]^{h\times w\times3}$ the policy consumes is mapped by a closed-form
photometric transform
\begin{equation}
\mathcal{E}_{N}(\mathbf{f})=\mathrm{clip}\!\left[\,g\,\big(\mathrm{sat}_{s}\mathbf{f}\big)^{\gamma}\odot\boldsymbol\tau+\sigma\boldsymbol\epsilon\,\right],
\quad \boldsymbol\epsilon\sim\mathcal{N}(0,1),
\label{eq:relight}
\end{equation}
with $\mathrm{sat}_s$ the saturation scaling, $\boldsymbol\tau$ a per-channel tint and the noise seed
fixed per frame. This is a photometric perturbation (gamma, gain, saturation, noise), not a captured night scene. We call it the \emph{night-style perturbation} and use $(\gamma,g,s,\sigma)=(2.2,0.45,0.45,7/255)$ with a blue tint (\cref{fig:method}a).

\paragraph{Validity of the interventions}
An independent detector (Faster R-CNN~\cite{ren2015faster}, COCO, score $\ge0.5$) recovers the target
in \NumDetOrig\% of the original images, in \NumDetStill\% under $\mathcal{E}_R$ and in
\NumDetNight\% under $\mathcal{E}_N$ (IoU $\ge0.5$): removal suppresses the pedestrian for an observer
outside the policy, re-lighting leaves it visible. Pasting inpainting patches on empty road bounds the
residual artefact at $5$--$17\%$ of the removal effect.

\subsection{Readouts}
With $c_t(X)=\|X_t-P_t\|-r_e-r_p$ the clearance net of the ego half-width $r_e=1.0$\,m and the
pedestrian radius $r_p=0.4$\,m, a plan is summarised by four functionals,
\begin{align}
S(X)&=\langle\min(c_t,\bar c)\rangle_t,\; A(X)=\mathbb{1}[\min_t c_t>0],\label{eq:SA}\\
C(X)&=\mathrm{clip}_{[0,1]}(\min_t c_t/d_0),\; T(X)=\inf\{t\!:c_t<d_0\}/2.5,\label{eq:CT}
\end{align}
with $T=1$ if the set is empty. The constants are physical, not tuned: $r_e$ is the half-width of the
nuScenes ego box ($4.08\times1.85$\,m), $r_p$ the disc a standing adult occupies, $d_0=1$\,m the outer
edge of personal space~\cite{hall1966hidden}, and $\bar c=10$\,m caps the contribution of scenes in
which the pedestrian is far away. Equations~\eqref{eq:SA}--\eqref{eq:CT} are the standard surrogate-safety measures~\cite{gettman2003surrogate}, read as distance to a person over the whole trajectory. Speed is deliberately not a readout: the same deceleration is safe or unsafe depending on where the person is.

\subsection{Exams}
Let $L(X)=\sum_t\|X_{t+\delta}-X_t\|$ denote arc length and $H$ the logged human trajectory.

\paragraph{Exposure}
$\mathrm{Exp}=\mathrm{med}\,L(X^{O})/L(H)$ over frames with $L(H)\ge0.5$\,m.

\paragraph{Hazard sensitivity}
A unit \emph{needs} a reaction when the pedestrian-blind plan enters personal space, $C(X^{R})<1$. On
those units, writing $\Delta(\cdot)=(\cdot)(X^{O})-(\cdot)(X^{R})$,
\begin{equation}
\HS=\tfrac14\big[\Delta A+\Delta C+\Delta T+\tanh(\Delta S/d_0)\big]\in[-1,1].
\label{eq:hs}
\end{equation}
Each term lies in $[-1,1]$ by construction and $\tanh$ squashes the single unbounded one at the $d_0$
scale. The composite is ours, its ingredients are not. Its four terms are the surrogate-safety
measures of \eqref{eq:SA}--\eqref{eq:CT}, and averaging normalised sub-scores into one index is how
the benchmark we compare against forms its extended predictive driver model score (EPDMS)~\cite{dauner2024navsim}, keeping the two commensurable.
What the check-up adds is that every sub-score is a \emph{paired difference} across
$\mathcal{E}_R$---the ablation form used to attribute behaviour to an input in interpretability
work~\cite{zou2023repe}. $\HS$ is read against the model-independent hazard
\begin{equation}
a_{\mathrm{req}}=\frac{v^{2}}{2\max(d-2,\,0.5)},\qquad \mathrm{TTC}_0=\frac{d-2}{v},
\label{eq:hazard}
\end{equation}
the deceleration required to stop short of a pedestrian at distance $d$ and the corresponding
time to collision (TTC)~\cite{gettman2003surrogate}. Units with $v<1$\,m/s are excluded: a
stationary ego can move neither toward nor away from anyone.

\paragraph{Scaling}
$\mathrm{Sc}=\rho\big(\HS,\,a_{\mathrm{req}}\big)$ over needed units, with $\rho$ Spearman's rank
correlation.
\paragraph{Specificity}
On units where the pedestrian requires no reaction ($C(X^{R})\ge1$) the correct plan does not change when it is removed, and a plan that changes anyway over-reacts:
\begin{equation}
\SP=\exp\!\big(-\bar D(X^{O},X^{R})/0.5\,\mathrm{m}\big)\in(0,1],
\label{eq:sp}
\end{equation}
so $\SP\to1$ iff removing a pedestrian that required no reaction leaves the plan where it was. $\SP$ is read together with $\HS$: a policy blind to pedestrians scores $\SP=1$ and $\HS=0$.

\paragraph{Lighting}
The \emph{causal faithfulness ratio} ($\CFR$) weighs the two displacements of \eqref{eq:FI} against
each other,
\begin{equation}
\CFR=\frac{\mathbb{E}\,\bar D(X^{O},X^{R})}{\mathbb{E}\,\bar D(X^{O},X^{N})}=\frac{\mathbb{E}F}{\mathbb{E}I},
\label{eq:cfr}
\end{equation}
A policy for which lighting is irrelevant has $\CFR\gg1$. The thresholds ($\HS\ge0.15$, $\mathrm{Sc}>0$, $\SP\ge0.8$, exposure within $[0.8,1.2]$, $\CFR>1$) are reference values we set, not population norms.

\section{Experiments}
\label{sec:exp}

\subsection{Setup}
\label{sec:setup}
We examine LTF and DiffusionDrive (front camera and ego status), DiffusionDriveV2 (camera and LiDAR),
SimLingo (InternVL2-1B), AutoVLA (Qwen2.5-VL-3B, three cameras $\times$ four frames) and Alpamayo-1.5 (four cameras, a diffusion head over the vision--language model cache), all with released weights. The three BEV planners are trained on NAVSIM (nuPlan) data, AutoVLA's released checkpoint on a mixture of nuPlan and nuScenes~\cite{zhou2025autovla}, SimLingo in CARLA, and Alpamayo-1.5 on NVIDIA's internal fleet data together with public sets that include NAVSIM-Traj and the nuScenes-derived nuScenesQA, DriveLM and OmniDrive~\cite{nvidia2026alpamayo15}. None is fine-tuned here. On the nuScenes frames the BEV planners and SimLingo are therefore out of distribution and the two VLA policies are not, and on the NAVSIM scenes only SimLingo is. $F<I$ on average for every policy (\cref{sec:axes}, \cref{sec:lighting}). For the BEV planners the nuScenes front image is centre-cropped to the 4:1 aspect of NAVSIM's stitched front view with the sky removed and resized to the network's input size, the logged speed and acceleration fill the status vector, and DiffusionDriveV2 receives the LiDAR sweep. The VLA policies receive the nuScenes cameras mapped approximately onto the views they expect. DiffusionDrive's diffusion seed is fixed so that paired queries share their noise.

The primary corpus is nuScenes~\cite{caesar2020nuscenes} keyframes: corridor pedestrians (lateral
clearance to the ego corridor $\le1$\,m) at $5$--$12$\,m, their off-road counterparts, and $30$
approach sequences at $5.8$--$14.6$\,m with logged pedestrian futures. The near-pedestrian units come from $88$ Boston (left-hand-drive vehicles, right-hand traffic) and $148$ Singapore (right-hand-drive vehicles, left-hand traffic) front-only frames. Alpamayo-1.5 covers $170$ of them, needing $1.6$\,s of history and $6.4$\,s of future
within the scene. \Cref{sec:axes} repeats $\mathcal{E}_R$ and $\mathcal{E}_N$ on a second, right-hand-drive corpus, the \NumSGscenes{} NAVSIM test-split~\cite{dauner2024navsim} Singapore scenes with a pedestrian near the ego corridor and a complete logged pedestrian future, used for the transfer and decomposition results and for the NAVSIM columns of \cref{tab:report}. The exam rows of \cref{tab:report} are measured on the nuScenes frames.

\subsection{The Two Axes}
\label{sec:axes}

The diagnosis reads a policy on the two axes of \eqref{eq:FI}. We apply both interventions to the
\NumSGscenes{} NAVSIM Singapore near-pedestrian scenes and read, per scene, $F$, the plan
displacement caused by removing the target pedestrian, and $I$, that caused by the night-style
perturbation.

\begin{figure}[t]
\centering
\includegraphics[width=\columnwidth]{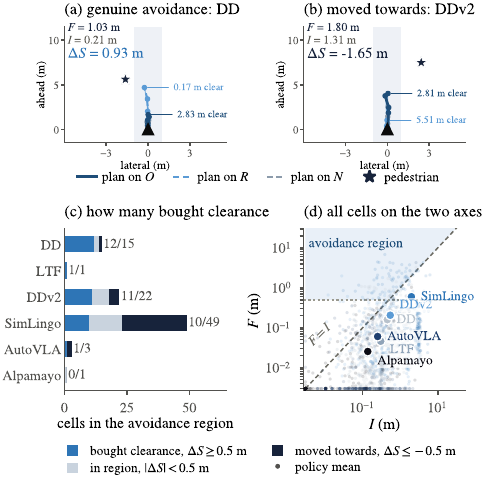}
\caption{\textbf{The two axes.} (a)~Genuine avoidance: with the pedestrian in view ($X^{O}$, solid)
DiffusionDrive stays short of its logged future (dashed from the star), while blind to it ($X^{R}$, dashed) it reaches it. $X^{N}$ gives $I$. (b)~Moved towards: DiffusionDriveV2 plans further with the
pedestrian than without. (c)~Cells in the avoidance region ($F\ge0.5$\,m, $F>I$) per policy, by $\Delta S$: blue $\ge0.5$\,m, navy $\le-0.5$\,m, grey between. Label: blue over all. (d)~All $1{,}476$ cells (\NumSGscenes{} scenes $\times$ six policies) on the axes of \eqref{eq:FI}, by policy, with means.}
\label{fig:cases}
\end{figure}

$F$ measures how far a plan moves, not where it moves to. We therefore decompose it (\cref{fig:cases}d), reading $F$ against $I$ of
\eqref{eq:FI}---the same policy's interference under an unrelated edit---and against $\Delta S$, the
change in clearance to the pedestrian's logged future. A response counts as avoidance only if the plan moved ($F\ge0.5$\,m) and the movement bought distance ($\Delta S\ge0.5$\,m), and, where a re-lit plan exists, only if it moved more than the policy's own interference ($F>I$).

Four readings follow. First, the pedestrian outweighs the irrelevant edit in only
\NumFgtIlo--\NumFgtIhi\% of scenes, so most of what $F$ registers sits below the interference of an irrelevant edit, the $F=I$ line of \cref{fig:cases}d. Second, avoidance is rare where it is called for. At the logged speed, where all three conditions apply, 35 of the $1{,}476$ policy--scene cells qualify (\cref{fig:cases}c). The logged ego is
stationary through most of this corpus, so a plan that does not move is usually the correct one. We therefore replay every scene at input speeds of 4 and 8\,m/s, as in \cref{sec:checkup}, and read only the cells in which the pedestrian-blind plan does come within 1\,m of the pedestrian's logged future, the same criterion as $C(X^{R})<1$ in \cref{sec:checkup}. No re-lit plan exists at these speeds, so the $F>I$ condition is not applied there.
Across \NumCfModels{} policies that is \NumCfNeed{} of \NumCfCells{} cells (\NumCfNeedLo--\NumCfNeedHi\%
per policy), and \NumCfReal{} of them---\NumCfRealPct\%---are genuine avoidance. On those cells the
median clearance changes by \NumCfDsLo--\NumCfDsHi\,m and the median planned distance by
\NumCfArcLo--\NumCfArcHi\,m: facing a pedestrian it would otherwise reach, the median plan neither diverts nor slows under our open-loop protocol. At the logged speed alone the need set has 53 cells and one genuine avoidance, so the counterfactual speeds add cells without changing the rate. Third, the displacement that does occur is not aimed: of the
\NumMoveN{} cells in which the plan moved, \NumMoveFar{} increased the clearance and \NumMoveNear{}
decreased it. Per policy the split differs sharply (\cref{fig:cases}c): DiffusionDrive reacts in few
scenes but in the right direction (median $\Delta S=\NumDDmed$\,m), while SimLingo moves in more
scenes than the other five combined and moves \emph{towards} the pedestrian more often than away
(\NumSLnear{} against \NumSLfar, median \NumSLmed\,m), and it also has the highest rate of sub-1.5\,s time-to-proximity (\cref{tab:report}). SimLingo is trained in simulation and run here on real images, so part of what we measure is a sim-to-real gap, which is exactly what a check-up at a new site should surface. Finally, what a policy yields does not grow with the hazard: the
rank correlation between $\Delta S$ and the deceleration the scene demands is between \NumRhoLo{} and \NumRhoHi{}.

An exam that rewarded responsiveness alone would therefore rank first the policy with the most sub-1.5\,s encounters. A score merges these four readings into one number. The edited frames keep them apart.

\subsection{The Five Exams}
\label{sec:diagnosis}

\begin{table*}[t]
\centering
\caption{\textbf{The full report for the six policies.} Rows are readings, columns policies. Brackets: 95\% CI. Arrows: outside the reference threshold of \cref{sec:checkup}. Bold: best per row ($p<0.01$ for the two rows of $N$). Human: the logged trajectory scored against each policy's blind plan. EPDMS on all 783 and on the \NumNclose{} near-pedestrian NAVSIM scenes.}
\label{tab:report}
\footnotesize
\setlength{\tabcolsep}{4pt}
\begin{tabular}{@{}lccccccc@{}}
\toprule
Reading\,$\backslash$\,Policy & DD & LTF & DDv2 & SimLingo & AutoVLA & Alpamayo 1.5 & Human \\
\midrule
Exposure (ratio to human) $\approx1$ & 0.50$\downarrow$ & 0.79$\downarrow$ & 0.88 & 2.19$\uparrow$ & \textbf{1.01} & 1.13 & 1.00 \\
\multicolumn{1}{r}{\textcolor{gray}{\scriptsize 95\% CI}} & \textcolor{gray}{\scriptsize[0.46, 0.55]} & \textcolor{gray}{\scriptsize[0.75, 0.84]} & \textcolor{gray}{\scriptsize[0.77, 0.95]} & \textcolor{gray}{\scriptsize[1.98, 2.65]} & \textcolor{gray}{\scriptsize[0.97, 1.11]} & \textcolor{gray}{\scriptsize[1.06, 1.21]} & \\[-1pt]
Hazard sens.\ $\HS\in[-1,1]$ $\uparrow$ & 0.06$\downarrow$ & 0.03$\downarrow$ & \textbf{0.10$\downarrow$} & 0.00$\downarrow$ & 0.02$\downarrow$ & 0.01$\downarrow$ & 0.57 \\
\multicolumn{1}{r}{\textcolor{gray}{\scriptsize 95\% CI}} & \textcolor{gray}{\scriptsize[$-$0.01, 0.14]} & \textcolor{gray}{\scriptsize[0.00, 0.06]} & \textcolor{gray}{\scriptsize[0.05, 0.16]} & \textcolor{gray}{\scriptsize[$-$0.03, 0.03]} & \textcolor{gray}{\scriptsize[0.00, 0.05]} & \textcolor{gray}{\scriptsize[$-$0.01, 0.02]} & \\[-1pt]
Scaling (rank corr.) $\uparrow$ & $-$0.28$\downarrow$ & $-$0.13$\downarrow$ & \textbf{0.23} & $-$0.22$\downarrow$ & $-$0.18$\downarrow$ & $-$0.23$\downarrow$ & 0.13 \\
\multicolumn{1}{r}{\textcolor{gray}{\scriptsize 95\% CI}} & \textcolor{gray}{\scriptsize[$-$0.50, $-$0.05]} & \textcolor{gray}{\scriptsize[$-$0.34, 0.07]} & \textcolor{gray}{\scriptsize[$-$0.02, 0.50]} & \textcolor{gray}{\scriptsize[$-$0.38, $-$0.08]} & \textcolor{gray}{\scriptsize[$-$0.34, 0.01]} & \textcolor{gray}{\scriptsize[$-$0.49, 0.03]} & \\[-1pt]
Specificity $\SP\in(0,1]$ $\uparrow$ & 0.76$\downarrow$ & 0.78$\downarrow$ & 0.65$\downarrow$ & 0.44$\downarrow$ & 0.76$\downarrow$ & \textbf{0.84} & -- \\
\multicolumn{1}{r}{\textcolor{gray}{\scriptsize 95\% CI}} & \textcolor{gray}{\scriptsize[0.73, 0.80]} & \textcolor{gray}{\scriptsize[0.75, 0.81]} & \textcolor{gray}{\scriptsize[0.61, 0.70]} & \textcolor{gray}{\scriptsize[0.38, 0.50]} & \textcolor{gray}{\scriptsize[0.72, 0.79]} & \textcolor{gray}{\scriptsize[0.81, 0.87]} & \\[-1pt]
Lighting $\CFR$ (ratio) $\uparrow$ & 0.23$\downarrow$ & 0.23$\downarrow$ & \textbf{0.47$\downarrow$} & 0.23$\downarrow$ & 0.44$\downarrow$ & 0.25$\downarrow$ & -- \\
\multicolumn{1}{r}{\textcolor{gray}{\scriptsize 95\% CI}} & \textcolor{gray}{\scriptsize[0.17, 0.30]} & \textcolor{gray}{\scriptsize[0.19, 0.28]} & \textcolor{gray}{\scriptsize[0.37, 0.58]} & \textcolor{gray}{\scriptsize[0.18, 0.29]} & \textcolor{gray}{\scriptsize[0.33, 0.55]} & \textcolor{gray}{\scriptsize[0.16, 0.35]} & \\[-1pt]
\midrule
\emph{Collision} at 8\,m/s, vis.\,/\,rm.\ (\%) $\downarrow$ & 7.6\,/\,9.2 & 9.9\,/\,10.7 & \textbf{3.1\,/\,6.1} & 40.5\,/\,36.6 & 32.1\,/\,29.8 & 11.2\,/\,11.2 & -- \\
\midrule
speed of $N$ (\%) $\downarrow$ & \textbf{\boldmath $37\%$} & \textbf{\boldmath $12\%$} & \textbf{\boldmath $-8\%$} & \textbf{\boldmath $16\%$} & $-0.3\%$ & $-2\%$ & -- \\
clearance of $N$, $\Delta_N S$ (m) $\uparrow$ & \textbf{\boldmath $-0.37$} & \textbf{\boldmath $-0.24$} & \textbf{\boldmath $0.15$} & \textbf{\boldmath $-0.47$} & $0.005$ & $0.03$ & -- \\
\midrule
\multicolumn{8}{@{}l}{\emph{Standard scores} (rank)} \\
L2 (m), vis.\,/\,rm. $\downarrow$ & 0.98\,/\,0.99 & 0.88\,/\,0.91 & 1.35\,/\,1.31 & 3.28\,/\,3.35 & 0.68\,/\,0.66 & 0.77\,/\,0.72 & -- \\
EPDMS, all scenes $\uparrow$ & 0.535 (4) & 0.509 (5) & 0.652 (3) & 0.359 (6) & \textbf{0.683 (1)} & 0.663 (2) & -- \\
EPDMS, near-ped. $\uparrow$ & 0.775 (2) & 0.774 (3) & 0.371 (6) & 0.617 (4) & 0.414 (5) & \textbf{0.788 (1)} & -- \\
TTC$<$1.5\,s (\%), LHD $\downarrow$ & \textbf{3.2 (1)} & 5.4 (2) & 18.3 (3) & 61.3 (6) & 19.4 (4) & 25.4 (5) & -- \\
TTC$<$1.5\,s (\%), RHD $\downarrow$ & \textbf{13.4 (1)} & 15.9 (2) & 22.0 (5) & 56.1 (6) & 20.7 (4) & 19.5 (3) & -- \\
\bottomrule
\end{tabular}
\end{table*}

\Cref{tab:report} gives the six profiles, read under our open-loop protocol on the nuScenes frames: the behaviour behind the ranks of \cref{sec:sample}.

\paragraph{Exposure}
The six policies differ first in how far they plan, and this alone will turn out to explain more of
the collision picture below than perception does: exposure spans a factor of four, from DiffusionDrive
at half the human's distance to SimLingo at more than twice it, with AutoVLA and Alpamayo-1.5 near the
human's own.

\paragraph{Hazard sensitivity}
On these frames the plans respond little to the pedestrian: on units that need a reaction, seeing it buys only \NumHSlo--\NumHShi{} of safety on a $[-1,1]$ scale, against \NumHumanRef{} when the logged human trajectory $H$ takes the place of $X^{O}$ in \eqref{eq:hs} (\cref{tab:report}), a reference rather than a ceiling, since $H$ and $X^{R}$ differ for reasons beyond the pedestrian. The benefit is also unreliable in direction:
counting only cases where seeing the pedestrian changes the outcome, it turns a collision into a clean
pass about as often as it turns a pass into a collision, for every policy (DiffusionDrive 9 vs.\ 5,
SimLingo 21 vs.\ 30, the rest at most 5 either way, all exact-binomial $p\ge0.26$) --- what reads as a safety margin in \cref{tab:report} does not carry over to the individual case.

\paragraph{Scaling}
The benefit does not grow with the hazard: its rank
correlation with $a_{\mathrm{req}}$ is negative for \NumScalNeg{} policies, down to \NumScalLo{} for
\NumScalWorst. DiffusionDriveV2 is the only exception (\NumScalHi), and its benefit stays below
$0.15$. Replacing the hazard axis by the time-based $\mathrm{TTC}_0$, whose sign is reversed by
construction, leaves the picture unchanged (\NumScalTtcLo{} to \NumScalTtcHi, against \NumScalLo{} to
\NumScalHi). The human reference is $\NumHumanSc$: a driver who already yields at every hazard level has little room left to scale, so this exam is informative where $\HS$ is small, which is where all six policies sit.

\paragraph{Specificity}
A policy whose plan moves when nothing requires it is not cautious but unsteady, and steadiness is what the exploratory analysis of \cref{sec:sample} finds to travel across driving sides. SimLingo's
plan moves most when nothing requires it ($\SP=\NumSpSL$), while the other five stay steadier, at
\NumSpOtherLo--\NumSpOtherHi.

\paragraph{Lighting}
Removing the pedestrian moves every plan less than the night-style perturbation does ($\CFR<1$ for all six, \cref{tab:report}). \Cref{sec:lighting} reads this exam in depth.

\paragraph{Collisions}
Collisions with the pedestrian's logged future follow how far a policy plans to drive, not what it perceives. At the logged
speed, the visible and removed rates differ by at most \NumCollGap{} points for every policy, and at
8\,m/s by at most \NumCollGapEight. SimLingo, which plans the longest trajectories, collides most. The low-exposure
policies almost never reach the pedestrian. DiffusionDrive is safe because it stays short, AutoVLA and Alpamayo-1.5 cover a human-like distance
without yielding, and SimLingo covers the most ground with the lowest specificity.

\subsection{The Lighting Exam in Depth}
\label{sec:lighting}

Lighting should not change what a driver does, so this exam asks how far a policy is moved by an
irrelevant change. It is the verdict that transfers between driving sides for every policy (\cref{sec:sample}).

\paragraph{Magnitude}
Removing a pedestrian from the corridor moves every policy's plan only \NumCFRlo--\NumCFRhi{} times as
far as the night-style perturbation does, with every 95\% interval below one (\cref{tab:report}). On detector-verified edits $\CFR=0.30$--$0.64$.

\paragraph{Direction: faster, and therefore closer}
Three policies plan faster under the night-style perturbation, DiffusionDrive by \NumNightDD\%,
SimLingo by \NumNightSL\% and LTF by \NumNightLTF\%, and their plans pass $0.24$--$0.47$\,m closer to
the pedestrian, while DiffusionDriveV2, the only policy that also reads LiDAR, slows
($\NumNightDDv\%$) and keeps more distance. Whether these profiles and orderings persist across driving sides is tested in \cref{sec:sample}.

\subsection{Prognosis: What Transfers, and at What Cost}
\label{sec:sample}

\begin{figure}[t]
\centering
\includegraphics[width=\columnwidth]{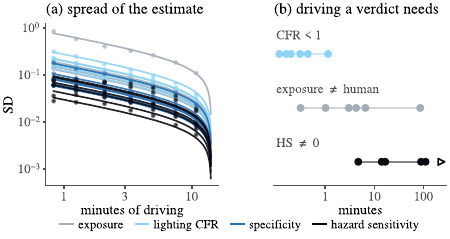}
\caption{\textbf{What a verdict costs.} One frame is 5\,s of near-pedestrian driving. (a)~Spread of
each estimate against sample size with its fitted law, one line per policy. (b)~Driving needed per
verdict, one dot per policy, off-scale values at the axis end.}
\label{fig:sample}
\end{figure}

Two questions decide whether a few hundred edited frames can stand in for a rank: which verdicts carry over to the other driving side, and how many frames each verdict needs.

\paragraph{A pre-registered transfer test}
Does a diagnosis made on one driving side describe the same policy on the other? Before computing any
right-hand-drive statistic, we fixed seven predictions, P1--P7 in \cref{tab:prereg}, and their decision rules. They were derived from the profile of the 88 Boston
frames, and we tested them on the near-pedestrian units of the 148 Singapore frames
(\cref{tab:prereg}). The registration covered the six policies public at that time. NVIDIA
subsequently released Alpamayo~1.5, which accepts navigation instructions, and we use it in place of
Alpamayo-R1 throughout this paper. Because both driving sides of Alpamayo~1.5 were measured together,
its numbers do not satisfy the registration discipline, so the table is evaluated on the five policies
that do. Retaining the replaced policy changes none of the seven verdicts.

\begin{table}[t]
\centering
\caption{\textbf{Pre-registered transfer test.} Predictions fixed on 88 Boston frames, tested on 148 Singapore frames, five policies (the replaced sixth changes no verdict).}
\label{tab:prereg}
\footnotesize
\setlength{\tabcolsep}{2.5pt}
\begin{tabular}{@{}lp{4.35cm}cl@{}}
\toprule
No. & Prediction & Holds & Evidence \\
\midrule
P1 & Lighting outweighs the pedestrian (CFR $<1$) for every policy & yes & max upper CI 0.70 \\
P2 & Hazard sensitivity stays below 0.15 (upper CI) & \textbf{no} & DD, DDv2 above \\
P3 & Exposure ordering transfers ($\rho\ge0.6$; SimLingo top, DD bottom) & yes & $\rho=0.90$ \\
P4 & Specificity ordering transfers ($\rho\ge0.6$; SimLingo lowest $\SP$) & yes & $\rho=0.60$ \\
P5 & Point values stay within the left-hand CI ($\ge$70\% of 15 cells) & \textbf{no} & 60\% \\
P6 & Night- and removal-induced plan changes more orthogonal, less shift between sides (auxiliary) & \textbf{no} & $\rho=-0.90$ (opposite) \\
P7 & SimLingo collides most, visible vs.\ removed within 3\,pts & yes & gap 2.7\,pts \\
\bottomrule
\end{tabular}
\end{table}

\paragraph{What transferred}
\NumPass{} of the seven predictions hold: the lighting verdict, the two orderings and the collision outcome. The night-style perturbation
outweighs the pedestrian in every policy. Exposure keeps its order ($\rho=0.90$), and so does specificity, with SimLingo the lowest, at the registered threshold ($\rho=0.60$). SimLingo collides most, with the visible and removed
rates within 3 points of each other.

\paragraph{What did not}
Three predictions fail.
\begin{itemize}
\item Point values drift: only 60\% of the model--exam cells stay inside their Boston interval.
\item In Singapore, DiffusionDrive and DiffusionDriveV2 show a small but non-zero hazard sensitivity ($\HS=0.09$ and $0.11$) whose upper confidence limits, $0.18$ and $0.20$, cross the registered $0.15$ line. The failure is one of precision rather than of sign: $\HS$ stays below $0.12$ on both sides for every policy, but 148 frames do not pin it below $0.15$.
\item P6, on the alignment between the night-induced and the removal-induced plan changes, predicted that policies with more orthogonal changes would shift less between sides. It failed ($\rho=-0.90$, opposite to the prediction). We no longer use this quantity and report its verdict for completeness.
\end{itemize}

\paragraph{Reading}
A small check-up can therefore report an ordering and a verdict for use in a new domain, but not a
calibrated value.

\paragraph{The standard scores on the same frames}
Running the standard scores on the same six policies (\cref{tab:report}) shows what a rank cannot see.
Removing the pedestrian changes none of the nuScenes open-loop numbers, the L2 error by at most \NumMaxdLtwo\,m and the frontal and pedestrian collision rates by at most \NumMaxdCol{} and \NumMaxdPed{} points, so the metric cannot tell
whether a policy saw the pedestrian, and its ordering follows how far a policy's planned distance
departs from the human's (\NumRhoLtwoExp{} with $|\text{exposure}-1|$). On NAVSIM the EPDMS column follows the official protocol on the Singapore part of the test split and stands in for the leaderboard here. Its ordering over all 783 Singapore scenes is unrelated to the one on the \NumNclose{} near-pedestrian scenes ($\rho=\NumRhoAllClose$). Its first, AutoVLA, falls to fifth and its third, DiffusionDriveV2, to
last, because the full-set score is carried by the map-compliance gates (\NumMapShare\% of the spread)
and none of the exams above is visible in it. Pedestrian TTC, by contrast, orders the policies the
same way on both driving sides ($\rho=\NumRhoTTC$) and coincides with exposure (\NumRhoTTCExpL{},
\NumRhoTTCExpR).

\paragraph{Which exam travels (exploratory)}
The target is how often each policy keeps clear of the pedestrian on the Singapore scenes, the share
of moving-ego scenes with minimum time-to-proximity $\ge1.5$\,s, the complement of the TTC column of
\cref{tab:report}. This target changes dataset and city as well as driving side, whereas the pre-registered test above stays within nuScenes and changes city and side, but not dataset or sensor rig. Each predictor is a six-policy ordering obtained
without that data, and we read the rank correlations as orderings, not as tests. The leaderboard score and the open-loop error agree with the target at $|\rho|\le\NumStdRhoMax$ (NAVSIM leaderboard \NumRhoLbKeep, nuScenes L2 \NumRhoLtwoKeep). Pedestrian-centred readouts that need no counterfactual edit do better: the open-loop pedestrian-collision rate on the nuScenes near-pedestrian frames at \NumRhoPedcolKeep{} (ties averaged) and the TTC column of \cref{tab:report} taken on the left-hand-drive side at \NumOtherSideTTC. The exams are not a substitute for such measurements. Of the five, read on the 88 Boston frames as a pre-deployment diagnosis would be, specificity matches the target best, $\rho=\NumSpBehTab$ (\NumSpLhdList), against $\NumExpBehTab$ for exposure and $\NumCfrBehTab$ for $\CFR$, where specificity and
$\CFR$ are higher-is-better and exposure's negative value says that policies that plan further keep
clear less often. What the exams add is the reason a policy orders where it does: displacement that
is not aimed away from the pedestrian is not protection, and a policy producing a lot of it comes
close to the pedestrian on both sides.

\paragraph{The selection report}
Read together, the profiles of \cref{tab:report} give each policy a prognosis under our open-loop
protocol. SimLingo plans furthest, moves its plan when a pedestrian requires no reaction
($\SP=\NumSpSL$), moves towards the pedestrian more often than away, and collides most on both sides.
Trained in simulation, it would need target-domain post-training before deployment, and the check-up prices
the re-check at a few dozen frames for exposure and specificity. AutoVLA and Alpamayo-1.5 keep a
human-like exposure and a steady plan but show no hazard response. Before deployment, post-training should add the response to the pedestrian, and only hundreds of frames with a genuine need to react can verify that it did. DiffusionDrive and LTF are safe because
they plan short, which holds only while exposure stays low, so any post-training that lengthens their
plans should be re-examined for hazard sensitivity. DiffusionDriveV2 is the only policy with a positive scaling estimate (\NumScalHi, interval crossing zero) and the only one that slows under the night-style perturbation, at the cost of the lowest specificity among the BEV planners.

\paragraph{What a verdict costs}
For every exam and policy we subsample $n$ of the \NumPool{} frames that all six policies share, 400
times, and fit the spread of the estimate with $\mathrm{sd}(n)=a\sqrt{1/n-1/M}$, $M=\NumPool$ (\cref{fig:sample}).

\paragraph{Well-fitting exams}
The law is the finite-population sampling form, so the fit is not a test. What it yields is the spread constant $a$, which converts into the number of frames a verdict needs. For exposure, the lighting ratio and specificity the fit is tight ($R^2\ge0.97$) and the verdicts settle inside the pool.
\begin{itemize}
\item \emph{Lighting outweighs the pedestrian} ($\CFR<1$) is settled for every policy within
      \NumNstarCFRhi{} frames.
\item \emph{Exposure differs from the human's} is settled within \NumNstarExpHi{} frames for the four
      policies whose exposure is far from one. For AutoVLA and Alpamayo-1.5 it would take hundreds to
      thousands of frames, which is itself a finding, since they plan a human-like distance.
\item \emph{Orderings} of the six policies stabilise quickly: exposure with 10 frames
      (\NumRankExpTen\% of subsamples reach $\rho\ge0.6$ with the full ordering) and specificity with
      40 (\NumRankSPforty\%).
\end{itemize}

\paragraph{Poorly fitting exams}
Hazard sensitivity behaves differently. Its fits are poor ($R^2$ as low as 0.26) wherever few units need a reaction. Deciding that $\HS\neq0$ needs 57--1{,}326 frames, an extrapolation beyond the \NumPool-frame pool for four policies, except for SimLingo, whose estimate sits on the threshold, and the ordering reaches \NumRankHSninety\% stability only at 90 frames.

The division is therefore clear. A few dozen frames, \NumScCFR{} to \NumScExp{} logged scenes, diagnose exposure, specificity and lighting. Whether a policy manages hazard needs hundreds of frames that contain a genuine need to react.

\section{Discussion}
\label{sec:discussion}

\paragraph{Limitations}
The exams are open-loop, replayed from logs and re-anchored to the logged pose. The night-style
perturbation is photometric, not captured. The counterfactual pairs are edited, not captured: 8.5\% of
removals leave a detectable residue, the inpainting artefact is 5--17\% of the effect, and the VLA
policies are edited only through their front camera. Alpamayo-1.5 covers 170 of the nuScenes frames,
SimLingo's 2.5\,s plans are extrapolated to NAVSIM's 4\,s (holding position instead gives EPDMS 0.47, still the lowest), and cross-policy correlations rest on six
policies.

\paragraph{Next steps}
The check-up is the first half of a benchmark-to-deployment loop. The paired data will next be captured
rather than constructed, the same scene with and without a pedestrian recorded on our vehicle and
re-rendered with 3D Gaussian splatting. The day--night pairs will move from a photometric transform to
captured or physically rendered lighting. And the VLA policies will be post-trained on target-domain
data and re-examined, so that the diagnosis measures what post-training recovers. The sample-size laws
above price that collection. Beyond these steps, we intend to refine and systematise the check-up into a standing procedure for evaluating and selecting learned driving policies, so that a deployment decision can rest on a diagnosis rather than on a rank.

\section{Conclusion}
A benchmark score reports an outcome, not the behaviour behind it, so a rank cannot be trusted to hold
at a new site. The check-up reads that behaviour from a few hundred logged frames through verified,
pixel-level counterfactual edits, reduced to two causal axes, $F$ and $I$, on which five exams are
built. Three results carry the paper. On the cells in which a pedestrian lies on the planned path, only \NumCfRealPct\% of responses are genuine avoidance and, in open loop, the median clearance change is at most \NumCfDsHi\,m, while the night-style perturbation moves every plan further than the pedestrian does. Diagnosed on one
driving side and tested on the other, the exposure and specificity orderings, the lighting verdict and the collision outcome transfer, while point values and the hazard-sensitivity verdict do not, and the profiles read as a selection report: which policy is safe because it plans short, which covers a human-like distance without yielding, and which is unsteady under a change that requires no reaction. Each verdict is priced in advance, lighting within
\NumNstarCFRhi{} frames and exposure within \NumNstarExpHi, so that, for the verdicts that transfer, each diagnosis is priced at tens of frames rather than a fleet test.

\ifanon\else
\section*{Acknowledgment}
This research was supported by the U.S. Department of Transportation (No.\ 69A3552348305) and by grants from NVIDIA utilizing NVIDIA RTX PRO 6000 Blackwell Max-Q Workstation Edition GPUs.
\fi
\bibliographystyle{IEEEtran}
\bibliography{main}
\end{document}